\documentclass[runningheads]{llncs}
\usepackage{graphicx}
\usepackage{amsmath,amssymb} 
\usepackage{color}
\usepackage{graphicx}
\usepackage{amsmath,amssymb} 
\usepackage{ifthen}
\usepackage{float}
\usepackage{mathtools}
\usepackage{multicol}
\usepackage{subfigure}
\usepackage{xspace}
\usepackage{multirow}
\usepackage{enumitem}
\usepackage{booktabs}
\usepackage[algo2e, ruled, vlined, linesnumbered]{algorithm2e}
\usepackage{xcolor}	
\usepackage{setspace}
\usepackage{fixltx2e}
\usepackage{stackengine}[2013-10-15]

\makeatletter
\renewcommand*{\@fnsymbol}[1]{\ensuremath{\ifcase#1\or *\or \dagger\or \ddagger\or
    \mathsection\or \mathparagraph\or \|\or **\or \dagger\dagger
    \or \ddagger\ddagger \else\@ctrerr\fi}}
\makeatother

\begin{document}
\title{Visual Relationship Prediction via\\ Label Clustering and Incorporation of\\ Depth Information} %
\titlerunning{Relationship Prediction via Clustering and Depth Info.}

\author{\small{Hsuan-Kung Yang\and 
An-Chieh Cheng\thanks{indicates equal contribution}\and
Kuan-Wei Ho\textsuperscript{*}\and
Tsu-Jui Fu\and
Chun-Yi Lee}}
\authorrunning{H-K, Yang et al.}

\institute{Elsa Lab, Department of Computer Science, National Tsing Hua University
\email{\tt \scriptsize{\{hellochick, anjiezheng, firewings89504, rayfu1996ozig, cylee\}@gapp.nthu.edu.tw}}\\
}

\maketitle 
\vspace{-5mm}
\begin{abstract}
In this paper, we investigate the use of an unsupervised label clustering technique and demonstrate that it enables substantial improvements in visual relationship prediction accuracy on the Person in Context (PIC) dataset.  We propose to group object labels with similar patterns of relationship distribution in the dataset into fewer categories.  Label clustering not only mitigates both the large classification space and class imbalance issues, but also potentially increases data samples for each clustered category.  We further propose to incorporate depth information as an additional feature into the instance segmentation model.  The  additional  depth  prediction  path  supplements the  relationship  prediction  model  in  a  way  that  bounding  boxes    or segmentation masks are unable to deliver. We have rigorously evaluated the proposed techniques and performed various ablation analysis to validate the benefits of them.

\end{abstract}

\vspace{-2em}

\section{Introduction}
This paper describes an effective methodology to perform relationship prediction for the Person in Context (PIC) dataset~\cite{pic}. In this dataset, the primary objective is to estimate human-centric relations, such as human-to-human and human-to-object relations (e.g., relative positions and activities). Different from the previous datasets (e.g., the Visual Genome dataset~\cite{krishna2017visual} and the Open Images dataset~\cite{krasin2016openimages}) which only concern the relations between different bounding boxes in an image, PIC focuses on the relations between different instance segmentation.  This is especially challenging for complex scenes containing various sizes of overlapped objects, as they require a series of procedures in order to accurately predict the instance segmentation masks as well as evaluate their relations.

There have been a number of research works proposed in recent years aiming at solving the visual relationship prediction tasks~\cite{neural-motifs,relation-networks,deep-structured-learning}.  The authors in~\cite{neural-motifs} introduced a sequential architecture called MotifNet for capturing the contextual information between the bounding boxes of the objects in an image.  The contextual information is then used to construct a graph for representing the relationships of the objects in the image. Relation network~\cite{relation-networks} embraces a lightweight object relation module for modeling the relations between the objects in an image via the use of their appearance and geometric features.  A deep structured model is proposed in~\cite{deep-structured-learning} for predicting visual relationships at both the feature and label levels.  While the above techniques have shown significant promise in predicting relations between the bounding boxes in an image~\cite{neural-motifs,relation-networks,deep-structured-learning}, several key issues remain unsolved and can be summarized in three aspects.  First, the size of the classification space of the possible relations in a dataset is typically large.  As relationship prediction is essentially a classification problem~\cite{neural-motifs,relation-networks}, the large classification space usually leads to insufficient samples for each class and thus difficulties in training a classification model.  Second, significant class imbalance exists in most of the datasets~\cite{krishna2017visual,krasin2016openimages,pic}.  The uneven distribution of data samples further exacerbates the above issue of insufficient training data, resulting in a serious drop in prediction accuracy for certain infrequent classes.  Third, to the best of our knowledge, the visual relationship prediction task based on instance segmentations has yet been well explored. As instance segmentation has attracted considerable attention in the past few years~\cite{li2016fully,dai2016instance}, this special relationship prediction task calls for an approach toward tackling the above issues.    

In order to address the first two issues mentioned above, we investigate the use of an unsupervised label clustering technique and demonstrate that it enables substantial improvements in both accuracy and training efficiency.  We observe that in human-centric datasets such as PIC, a number of human-object pairs share similar patterns of relationship distribution, leaving significant opportunities for label clustering and data augmentation.  For example, small objects such as `bottle', `cellphone', and `plate' are extremely likely to have a `hold' relation with human beings.  Motivated by this observation, we investigate the use of k-means clustering technique to group object labels with similar patterns of relationship distribution in the dataset into fewer categories.  Label clustering not only mitigates both the large classification space and class imbalance issues, but also potentially increases data samples for each clustered category.  This provides the relationship prediction model more opportunities to improve its accuracy.  

To further enhance the relationship prediction model based on instance semantic segmentation, we additionally explore the use of depth information in relationship prediction.  We propose to incorporate depth information as an additional feature into the instance segmentation model. We observe that depth information is crucial in determining the spatial relations between objects such as `in-front-of', `next-to', `behind', and so on.  As a result, we integrate into our backbone instance segmentation model~\cite{mask-rcnn} with an auxiliary depth prediction path~\cite{monodepth}. The additional depth prediction path supplements the relationship prediction model in a way that bounding boxes ~\cite{neural-motifs,relation-networks,deep-structured-learning} or segmentation masks are unable to deliver.  This is due to the fact that they lack the three-dimensional information required for determining the spatial relations between objects.  We have rigorously evaluated the proposed techniques and performed various ablation analysis to demonstrate the benefits of them.  \\ \\
\\
The contributions of this paper include the following:
\begin{itemize}
\item A demonstration of visual relationship prediction based on semantic information extracted by an instance semantic segmentation model.
\item A method considering both the model architecture and data distribution.

\item A simple approach for dealing with the large classification space and class imbalance issues by clustering labels with similar relationship distribution.

\item An investigation into a novel concept of applying depth information into the relationship prediction task so as to provide additional spatial information.

\end{itemize}

The rest of this paper is organized as follows.  Section~\ref{sec::background} introduces background material.  Section~\ref{sec::methodology} walks through the proposed methodology, its implementation details, and the training procedure.  Section~\ref{sec::experiments} presents the experimental results and an ablation study of the proposed method.  Section~\ref{sec::conclusion} concludes this paper.

\section{Background}
\label{sec::background}
In this section, we introduce the knowledge background regarding relation prediction. We first provide an overview of instance segmentation. Then, we briefly review related works that focus on visual relation prediction. Finally, we introduce depth prediction which is applied in our method.

\subsection{Instance Segmentation}
Inspired by R-CNN \cite{girshick2014rcnn} and Faster R-CNN \cite{faster-rcnn}, many approaches to instance segmentation are based on segment proposals. DeepMask \cite{deepmask} learns to produce segment candidates and then do classification. Dai \textit{et al.} \cite{dai2016instance} presented a multi-stage cascade that predicts segment proposals based on bounding-box. In those methods, segmentation followed by recognition, which is slow and less accurate. Recently, FCIS \cite{li2016fully} combines segment proposal and object detection for fully convolutional instance segmentation. They predict different position-based output channels which are respectively for object class, bounding-box, and segmentation in fully convolutional way. In contrast to segmentation-first strategy, Mask R-CNN \cite{mask-rcnn} is an instance-first one. Based on Faster R-CNN, Mask R-CNN adds a new branch for object segmentation. They adopt two-stage process, with a Region Proposal Network (RPN) at first, followed by a sub-network which predicts object class and segmentation. In our proposed method, we apply Mask R-CNN, which is also a state-of-the-art network, as our instance segmentation network.

\subsection{Visual Relation}
Visual relation not only extracts objects' ground region but also describes their interactions. Yao \textit{et al.} \cite{Yao10a} first considers relation as hidden variables. Nowadays, there are several explicit relation extraction methods which can be divided into two categories: joint and separate. Joint methods ~\cite{ramanathanLDHLG15,VisualPhrases}
usually consider relation triplet as an unique class and generate class and relation information together. However, joint methods are faced with class imbalance problem and easily dominated by major class-relation triplet, which causes low accurate for minor ones. Conversely, separate models ~\cite{sadeghi2015viske,lu2016visual} first detect objects, and extract their relations individually.

Zhu \textit{et al.} \cite{liangLX17} presented a separate way for extracting visual relation. They applied Faster R-CNN for object detection and then feed object visual features into following CNN blocks to extract relation for each pair of object. Furthermore, Neural-Motifs \cite{neural-motifs} adopted two-stage bidirectional LSTM to model both global context and pair-wise object relation. The first stage of the network is for object recognition and the second stage is for relation extraction. In our baseline approach, we feed in visual features from Mask R-CNN and apply the second stage of Neural-Motifs as our relation extraction model.

\subsection{Depth Prediction}
Depth estimation from images has a long history in computer vision. In PIC challenge, we do not have scene geometry or other information for an image, so we only consider monocular depth estimation methods. Make3D \cite{make3d} over-segments input image into patches and then estimates the 3D location and orientation of local planes to explain each patch. Based on Make3D, Liu \textit{et al.} \cite{Liu:2016:LDS:3026801.3026841} use a CNN model to learn the global context and help generate more realistic output. Karsch \textit{et al.} \cite{karschLK14} further produces more consistent prediction via copying parts of depth images from training set.

Different from supervised-based monocular depth prediction, MonoDepth \cite{monodepth17} presents a unsupervised method. The author of MonoDepth train an FCN end-end to predict the pixel-level correspondence between pairs of stereo images and can perform single image depth estimation during testing. In our paper, we use the depth estimation from MonoDepth which is trained on Cityscapes \cite{Cordts2016Cityscapes} dataset as the auxiliary information for visual relation extraction.

\section{Methodology}
\label{sec::methodology}

\begin{figure}[tp!]
\centering
{\includegraphics[width=1.00\textwidth]{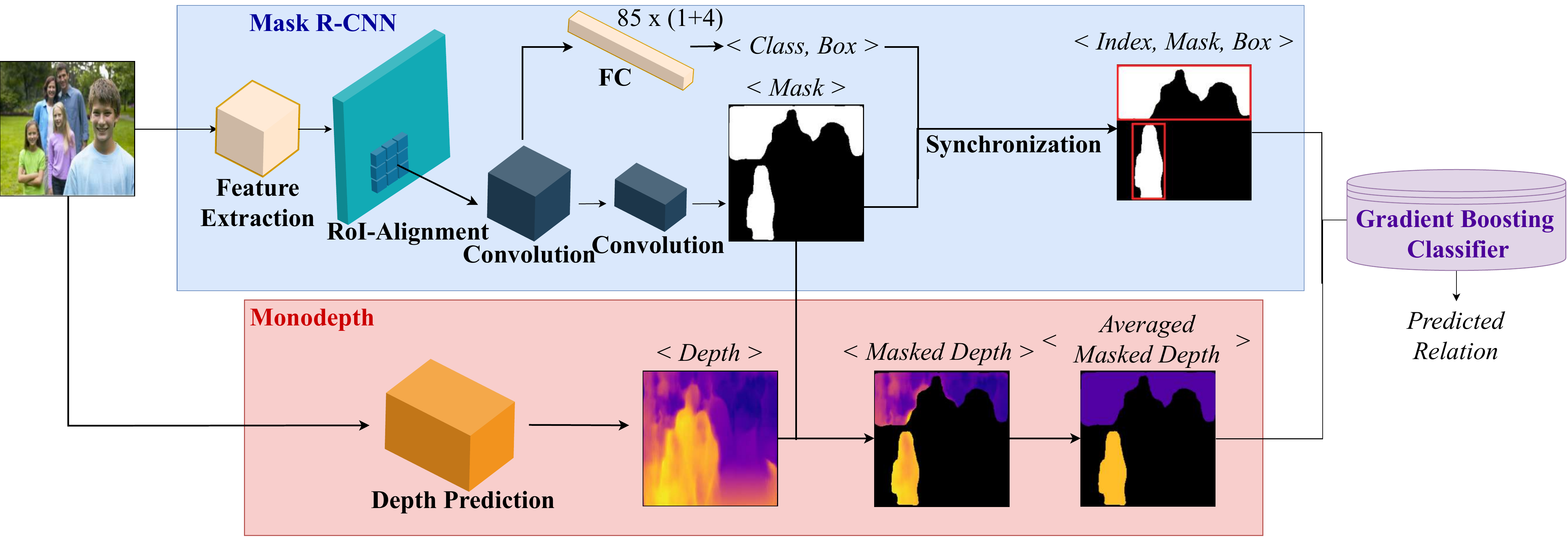}}
\caption{The proposed framework for relationship prediction.}
\vspace{-2em}
\label{fig:diagram}
\end{figure}

\subsection{Architecture Overview}
\label{subsec::architecture_overview}
Fig.~\ref{fig:diagram} illustrates the proposed architecture.  It consists of three components: an instance segmentation network (blue part), a depth prediction network (red part), and a gradient boosting classifier (purple part). The instance segmentation network is based on the implementation of Mask R-CNN~\cite{mask-rcnn}, which includes feature extraction layers, a region of interest (RoI) alignment layer, two convolutional layers for mask prediction, and a fully-connected (FC) layer for predicting class labels and bounding boxes for the objects contained in the figure.  A special synchronization procedure is used to align the indices of the bounding box and mask for each object in the figure.  Please note that the class labels here stand for clustered labels, which is explained in detail in Section~\ref{subsec::clustering}.  The depth prediction network is based on the implementation of Monodepth~\cite{monodepth}, which evaluates a depth map for the input image.  The depth map is then averaged within the mask region predicted by the instance segmentation network to generate an averaged masked depth map for the corresponding object instance.  Finally, the bounding boxes, masks, and the averaged masked depth maps are all fed into the gradient boosting classifier to predict relations between the objects in the input image.     

\subsection{Instance Segmentation Network}
We use Mask R-CNN~\cite{mask-rcnn} as our instance segmentation network. Mask R-CNN employs a two-stage structure, which includes a region proposal network (RPN) followed by a network consisting of a classification branch and a mask branch.  As plotted in Fig~\ref{fig:diagram}, the classification branch is used to predict classification scores and bounding boxes for the objects contained in the image.  On the other hand, the mask branch predicts an instance segmentation masks for each object.  As the indices of the objects from the two branches are different, a synchronization procedure is necessary so as to match the bounding boxes and masks.

\subsection{Depth Prediction Network}
We use Monodepth~\cite{monodepth} to implement the depth prediction network.  The network serves as an auxiliary path to the instance segmentation network and evaluates a depth map for each input image.  As illustrated in Fig~\ref{fig:diagram} and discussed in Section~\ref{subsec::architecture_overview}, the output of the depth prediction network is an averaged masked depth map.  This averaged masked depth map is generated from a portion of the pixels within the masked region contained in the raw masked depth map. As the raw depth map generated by the depth prediction network is not sufficiently accurate for all the pixels within the region of the mask, the depth value obtained by directly averaging these pixels is not representative of the masked object. As a result, we filter out the first and fourth quartiles of the depth values within the masked region.  The depth values of the remaining pixels are then averaged to generated the averaged masked depth map.  

\subsection{Clustering}
\label{subsec::clustering}
We reduce the 85 human-object categories by implementing unsupervised clustering on 85 human-object categories based on their relation's frequency distribution. The unsupervised clustering approach is based on the thought that “multiple objects might share the same frequency distribution of relations”. The clustering in advance helps model to reduce the computation space when outputting relation label as classification and augment categories with less data by clustering them with other categories to accumulate data at the same time. 

We choose K-means clustering as our clustering algorithm. K-means provides a simple strategy to cluster vectors quickly in a neat way. Our frequency distribution of 85 human-object categories will be normalized before sent into K-means. Moreover, we evaluate our unsupervised clustering with custom constraints. Because K-means require specific number of clusters $(n=k)$ as parameter, we propose a few constraints to determine whether the clustered result is best for the training and search for the optimal interval of $n$. The clustering evaluation is conducted under three constraints:
\begin{enumerate}
\item Objects inside the same group are similar enough with each other regarding frequency distribution
\item The number of clusters is small enough to benefit model’s computation
\item The total number of data inside each clusters are expected to be maximized
\end{enumerate}

The number of clusters most suitable under the three constraints lies in the range of 8 clusters to 10 clusters. The clustering curtails the original classification space (85 in original) in a significant range (8 after clustering). The optimal choice of number of clusters might vary depending on the attributes and distribution of the data.

\subsection{Relation Prediction}
In the PIC Challenge, every relations is human-centered. In other words, every relation candidates is predicted based on the previously generated pairs of each human in the instance and the rest of other objects. Say there is n instances in the image, and k human among the instances. In this case, the number of relation candidates will be $(n-1)*k$. 

The subject in each relation prediction must be human, so we only need to check out which pseudo label groups the object is to make decision. If action relations (non-geometric relation) are more frequent or serious class imbalance happens in the group, frequency based prediction will be adopted. The frequency based prediction method is a probability model whose every prediction is totally based on the distribution of each relation's frequency. That is to say, the more frequent a relation is the more probability it is predicted as answer. We also have the other form of frequency based method where only the most frequent relation label is predicted, discarding the rest of other less frequent relations. On the other hand, if geometric relations (especially 'in-front-of', 'next-to' and 'behind') are the major composition of the group's relation distribution, we will take the previously taken features (total 10 values) into training:

\begin{enumerate}
\item The bounding box difference (y1, x1, y2, x2) - 4 values
\item The bounding box overlap between subject and object
\item The pseudo label group of the object
\item The depth means and medians of subject and object - 4 values
\end{enumerate}

Those features will then be sent to gradient boosting classifier to train for the final relation prediction.

\section{Experimental Results}
\label{sec::experiments}

We have tried several different approaches on the PIC dataset. In this section, we will shed light on the big picture of our clustered results, and then we will break into the details of the approaches we mainly propose. Finally, we will show and compare every approaches we've experimented in an ablative analysis way.

\begin{figure}[tp!]
\centering
\subfigure[Before Clustering]
{\includegraphics[width=0.37\textwidth]{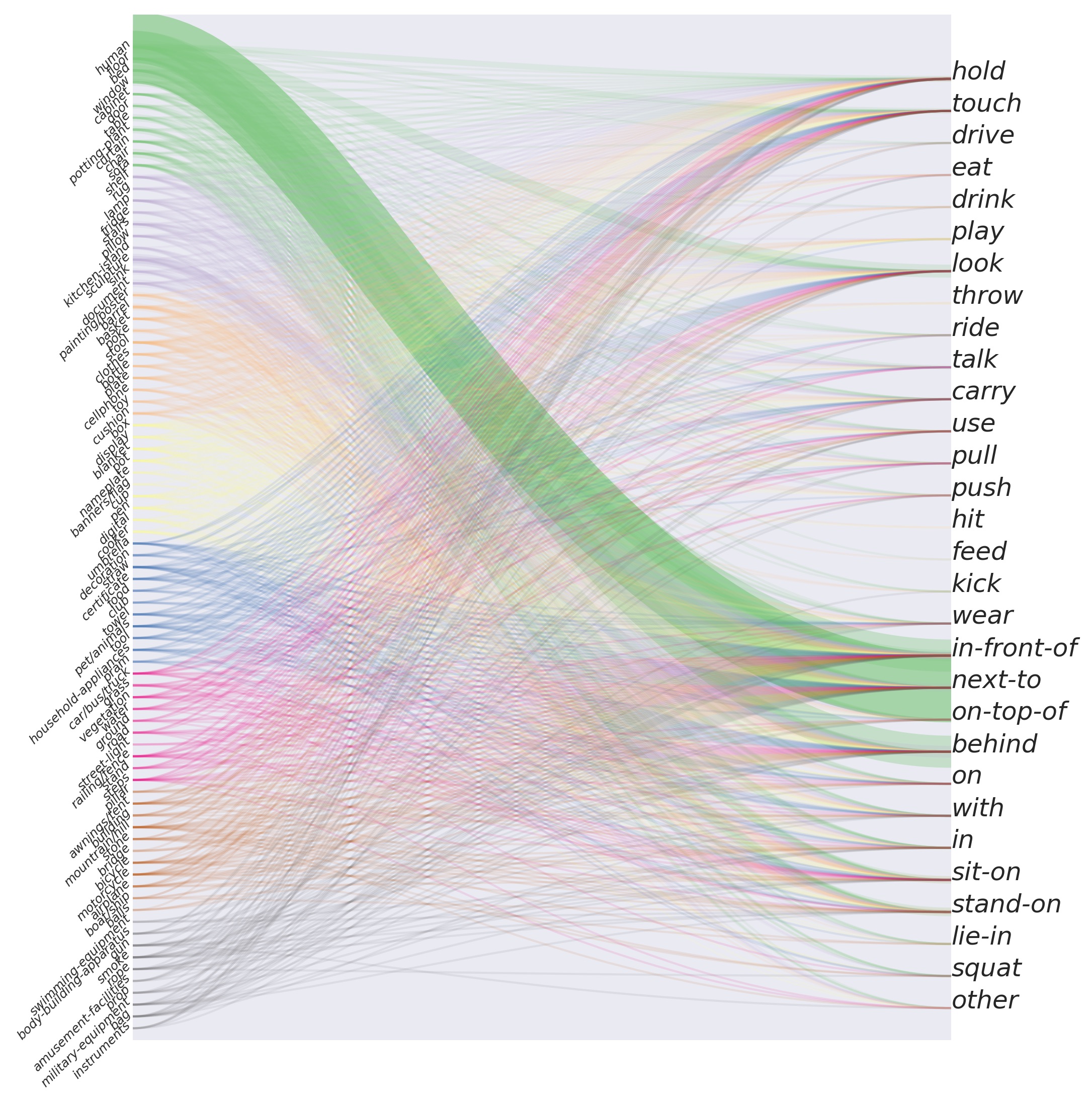}
\label{fig:beforeCluster}}
\subfigure[After Clustering]
{\includegraphics[width=0.37\textwidth]{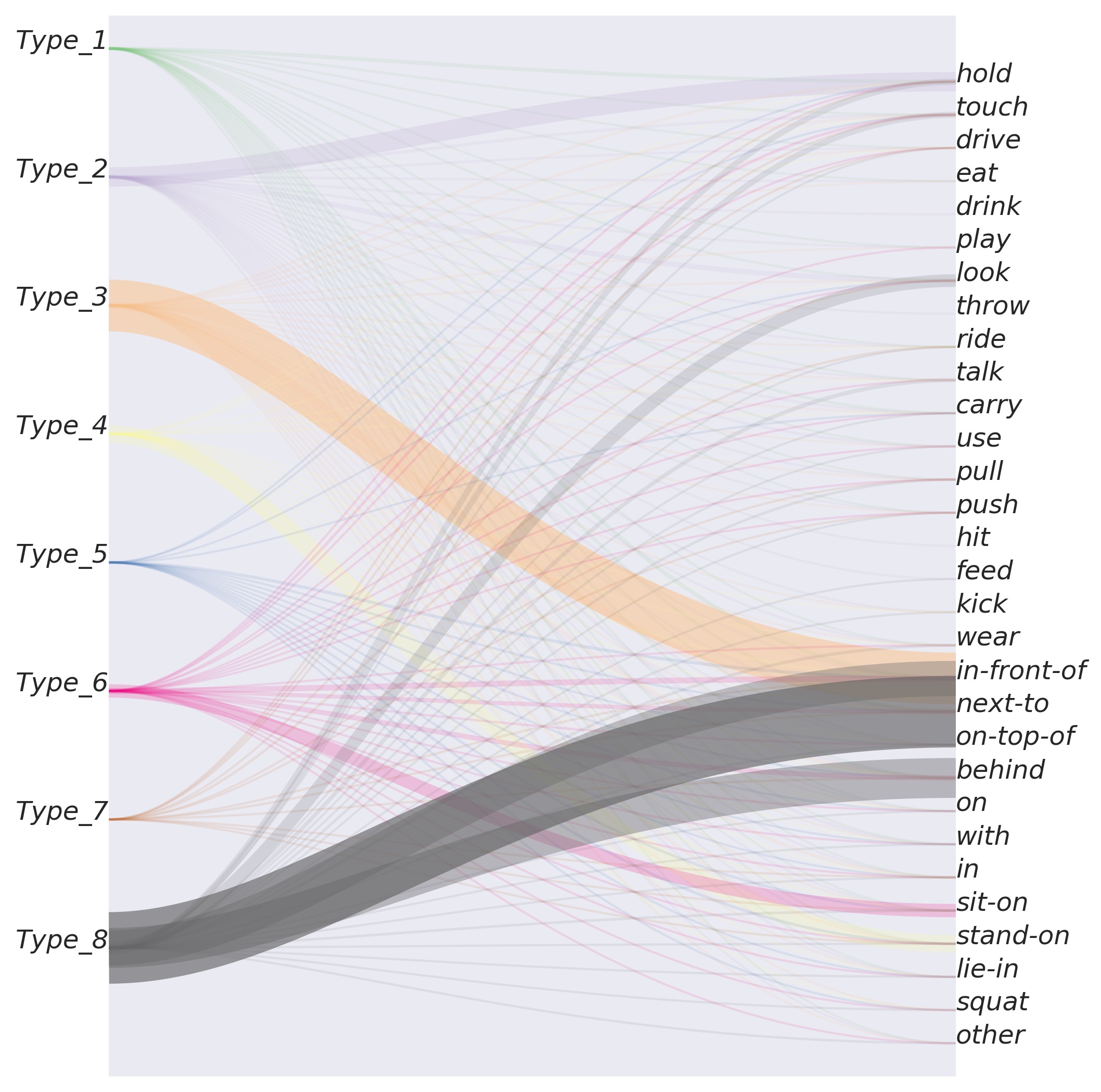}
\label{fig:afterCluster}}
\caption{Our proposed method clusters the 85 human-object categories into several pseudo label groups.\vspace{5pt}}
\vspace{-2em}
\label{fig:clusterablation}
\end{figure}

Before we dive into the details of our relation prediction method, we will introduce the result after clustering human-object categories into several pseudo label groups. 

The clustering results is shown in Fig. 2. Before the clustering, although there are 85 categories, only few category actually possesses significant relation numbers for the training, such as 'human'. After the clustering, we still can't say that the total numbers of each cluster are average enough, but the pseudo label group do augment the data similar in frequency distribution together. 
We hope the clustered result to meet the three constraints we proposed above as possible as they can.

We evaluate the similarity within each cluster by the standard deviation of every cosine distance between every frequency normalized vectors and the mean vectors. Finally, we pick the number of clusters as 8, which is the optimal choice we've tried so far. The clustered result is also shown in Fig. 2. 

We attempted several different approaches on relation prediction part.  First, we build the neural motifs~\cite{neural-motifs} model on the dataset as the baseline, and the instance segmentation model in this case we choose is Mask R-CNN. However, suffering from the large classification space and the class imbalance issues, neural motifs can't reach its full potential on the dataset. At first, we build a totally frequency based probability model to establish another baseline. Then we conduct experiments on the relation prediction we mentioned above, including the the prediction path decision regarding using frequency based prediction or gradient boosting. We also train the gradient boosting model with or without the mean and median of each instance depth to compare the performance.

\begin{figure}[tp!]
\centering
{\includegraphics[width=0.5\textwidth]{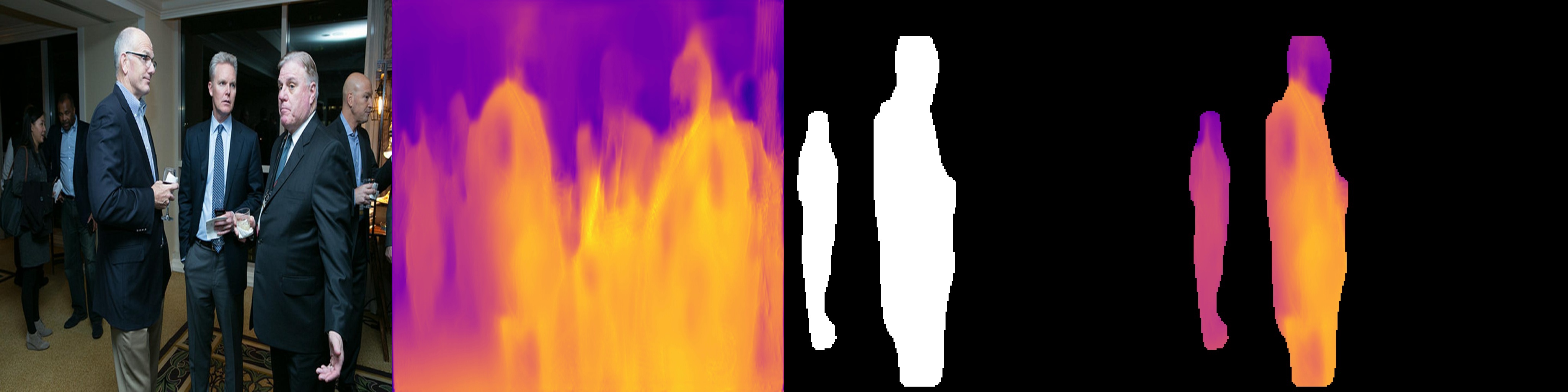}}
{\includegraphics[width=0.5\textwidth]{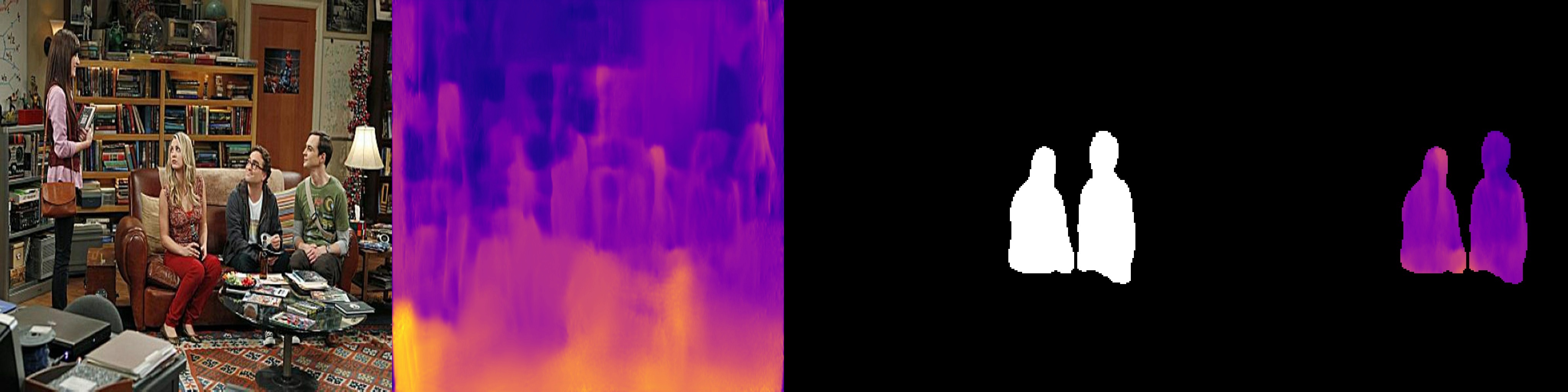}}
\caption{The illustration of masked depth. 
Those images from the 1st column to 4th column are raw images, depth images extracted from the raw image, instance segmentation of the images and the combined masked depth.}
\vspace{-1em}
\label{fig:depth}
\end{figure}

\begin{table}[h]
\centering
\resizebox{0.7\columnwidth}{!}{
\begin{tabular}{@{}c|c|c|c|c@{}}
\toprule
\textbf{Method} & \textbf{IoU(0.25)} & \textbf{IoU(0.5)} & \textbf{IoU(0.75)} & \textbf{Average}       \\ \midrule
Baseline        & 0.156 / 0.139      & 0.137 / 0.125     & 0.089 / 0.083      & 0.127 / 0.116          \\
Freq-85         & 0.312 / -          & 0.242 / -         & 0.133 / -          & \textbf{0.229 / -}     \\
Freq-8          & 0.317 / 0.186      & 0.246 / 0.151     & 0.134 / 0.082      & \textbf{0.232 / 0.140} \\
Gb              & 0.323 / -          & 0.256 / -         & 0.140 / -          & \textbf{0.240 / -}     \\
Gb-Depth        & 0.324 / 0.314      & 0.258 / 0.251     & 0.140 / 0.131      & \textbf{0.241 / 0.232} \\ \bottomrule
\end{tabular}
}
\vspace{5pt}
\caption
{ \small
'Baseline' in the method means Neural-Motifs; 'Freq-85' means frequency-based method on 85 categories; 'Freq-8' means 8 pseudo type; 'Gb' means gradient boosting; 'Gb-Depth' means with the depth information. The two scores under the same column are the validation score and test score. In PIC Challenge, the time of submission is limited, so not every method can be tested.
}
\vspace{-2em}
\end{table}

The results of each approaches are shown in the table. The score metric $IoU@n$ means that only predictions with $IoU$ metrics above threshold will be computed its relation accuracy. We can see the improvement from freq-85 to freq-8 in the table. The improvement indicates that clustering do improve the performance. Second, from the 3rd column to the 4th column, the performance is improved because of the additional depth information.

\section{Conclusions}
\label{sec::conclusion}

In this paper, we showed that unsupervised clustering can be effective in mitigating the large classification space and class imbalance issues in visual relationship prediction tasks.  We proposed a technique to cluster object labels of similar relationship distribution with human beings in the same dataset into categories.  In order to enhance the accuracy of instance segmentation based relationship prediction tasks, we further proposed to incorporate an auxiliary depth prediction path into our instance segmentation model. We demonstrated the effectiveness of the proposed techniques on the PIC dataset, with a detailed ablation study.

%
%
%

\begin{thebibliography}{9}
\footnotesize
\setlength{\itemsep}{0pt}

\bibitem{sadeghi2015viske}
S. M. Amin, F. Ali, ``VisKE: Visual Knowledge Extraction and Question Answering by Visual Verification of Relation Phrases," in {\it Proc. IEEE Conf. Computer Vision and Pattern Recognition (CVPR)}, pp. 1456--1464, Jun. 2015.


\bibitem{VisualPhrases}
S. M. Amin, F. Ali, ``Recognition using Visual Phrases," in {\it Proc. IEEE Conf. Computer Vision and Pattern Recognition (CVPR)}, pp. 1745-1752, Jun. 2011.


\bibitem{ramanathanLDHLG15}
Ramanathan et al., ``Learning semantic relationships for better action retrieval in images," in {\it Proc. IEEE Conf. Computer Vision and Pattern Recognition (CVPR)}, pp. 1100-1109, Jun. 2015.

\bibitem{Yao10a}
B. Yao, F. Li, ``Modeling Mutual Context of Object and Human Pose in Human-Object Interaction Activities," in {\it Proc. IEEE Conf. Computer Vision and Pattern Recognition (CVPR)}, 17-24, Jun. 2010.


\bibitem{li2016fully}
Y. Li, H. Qi, J. Dai, X. Ji, and Y. Wei, ``Fully Convolutional Instance-aware Semantic Segmentation," in {\it Proc. IEEE Conf. Computer Vision and Pattern Recognition (CVPR)}, pp. 4438-4446, Jul. 2017.

\bibitem{krishna2017visual}
R. Krishna {\it et al.}, ``Visual Genome: Connecting language and vision using crowdsourced dense image annotations," {\it Int. J. of Computer Vision (IJCV)}, vol. 123, pp. 32-73, May 2017.

\bibitem{krasin2016openimages}
I. Krasin {\it et al.}, ``OpenImages: A public dataset for large-scale multi-label and multi-class image classification," 2017. Dataset available from https://storage.googleapis.com/openimages/web/index.html.

\bibitem{pic}
`` PIC - Person In Context," Dataset available from http://picdataset.com/challenge/index/ .

\bibitem{neural-motifs}
R. Zellers, M. Yatskar, S. Thomson, and Y. Choi, ``Neural Motifs: Scene graph parsing with global context," in {\it Proc. IEEE Conf. Computer Vision and Pattern Recognition (CVPR)}, pp. 5831-5840, Jun. 2018.

\bibitem{relation-networks}
H. Hu, J. Gu, Z. Zhang, J. Dai, and Y. Wei, ``Relation networks for object detection," in {\it Proc. IEEE Conf. Computer Vision and Pattern Recognition (CVPR)}, pp. 3588-3597, Jun. 2018.

\bibitem{deep-structured-learning}
Y. Zhu and S. Jiang, ``Deep structured learning for visual relationship detection," in {\it Proc. AAAI Conf. Artificial Intelligence (AAAI)}, pp. 7623-7630, Feb. 2018.

\bibitem{faster-rcnn}
S. Ren, K. He, R. Girshick, and J. Sun, ``Faster R-CNN: Towards real-time object detection with region proposal networks," {\it IEEE Trans. Pattern Analysis and Machine Intelligence (TPAMI)}, vol 39, pp. 1137-1149, Jun. 2017.

\bibitem{monodepth}
C. Godard, O. M. Aodha, and G. J. Brostow, ``Unsupervised monocular depth estimation with left-right consistency," in {\it Proc. IEEE Conf. Computer Vision and Pattern Recognition (CVPR)}, pp. 270-279, Jul. 2017.

\bibitem{mask-rcnn}
K. He, G. Gkioxari, P. Dollár, and R. Girshick, ``Mask R-CNN," in {\it Proc. IEEE International Conf. Computer Vision (ICCV)}, pp. 2961-2969, Oct. 2017.

\bibitem{girshick2014rcnn}
R. Girshick, J. Donahue, T. Darrell, and J. Malik, ``Rich feature hierarchies for accurate object detection and semantic segmentation," in {\it Proc. IEEE Conf. Computer Vision and Pattern Recognition (CVPR)}, pp. 580-587, Jun. 2014.

\bibitem{deepmask}
P. O. Pedro, R. Collobert, and P. Dollar, ``Learning to segment object candidates," in {\it Proc. Int. Conf. Neural Information Processing Systems (NIPS)}, pp. 1990-1998, Dec. 2015.

\bibitem{dai2016instance}
J. Dai, K. He, and J. Sun, ``Instance-aware semantic segmentation via multi-task network cascades," in {\it Proc. IEEE Conf. Computer Vision and Pattern Recognition (CVPR)}, pp. 3150-3158, Jun. 2016.

\bibitem{sadeghi2015viske}
S. Fereshteh, D. Santosh K, and F. Ali, ``VisKE: Visual Knowledge Extraction and Question Answering by Visual Verification of Relation Phrases," in {\it Proceedings of the IEEE Conference on Computer Vision and Pattern Recognition}, pp. 1456-1464, May 2015

\bibitem{lu2016visual}
S. Fereshteh, D. Santosh K, and F. Ali, ``Visual Relationship Detection with Language Priors," in {\it Proceedings of the IEEE Conference on Computer Vision and Pattern Recognition}, pp. 852-869, 2016

\bibitem{liangLX17}
X. Liang, L. Lee, and E. Xing, ``Deep Variation-Structured Reinforcement Learning for Visual Relationship and Attribute Detection," in {\it 2017 {IEEE} Conference on Computer Vision and Pattern Recognition, {CVPR}}, pp. 4408-4417, 2017

\bibitem{make3d}
M. Sun and A. Y. Ng and A. Saxena, ``Make3D: Learning 3D Scene Structure from a Single Still Image," in {\it IEEE Transactions on Pattern Analysis \& Machine Intelligence}, pp. 824-840, May 2008

\bibitem{Liu:2016:LDS:3026801.3026841}
F. Liu, C. Shen, L. Guosheng, and I. Reid  ``Learning Depth from Single Monocular Images Using Deep Convolutional Neural Fields," in {\it IEEE Trans. Pattern Anal. Mach. Intell.}, pp. 2024-2039, Oct. 2016

\bibitem{karschLK14}
K. Kevin, L. Ce, and K. Sing Bing ``Depth Extraction from Video Using Non-parametric Sampling," in {\it Proc. Int. Conf. Neural Information Processing Systems (NIPS)}, pp. 2144-2158, May 2017.

\bibitem{monodepth17}
C. Godard, O. {Mac Aodha}, and G. J. Brostow, ``Unsupervised Monocular Depth Estimation with Left-Right Consistency," in{\it Proc. IEEE Conf. Computer Vision and Pattern Recognition (CVPR)}, pp. 6602-6611, Nov. 2017.

\bibitem{Cordts2016Cityscapes}
Marius et al., ``The Cityscapes Dataset for Semantic Urban Scene Understanding," in {\it Proc. of the IEEE Conference on Computer Vision and Pattern Recognition (CVPR)}, pp. 3213-3223, Dec. 2017.  

\end{thebibliography}
%

\end{document}